\documentclass[letterpaper, 10 pt, conference]{ieeeconf}  

\IEEEoverridecommandlockouts                              

\usepackage{amsmath,amssymb,amsfonts}
\usepackage{algorithmic}
\usepackage{graphicx}
\usepackage{textcomp}
\usepackage{xcolor}
\usepackage[algo2e,linesnumbered,lined,boxed,vlined]{algorithm2e}
    
\def\b{\ensuremath\boldsymbol}

\title{\LARGE \bf
Generative Locally Linear Embedding
}

\author{Benyamin Ghojogh$^{1}$, 
Ali Ghodsi$^{2}$, Fakhri Karray$^{3}$,~\IEEEmembership{Fellow,~IEEE},
Mark Crowley$^{1}$
\thanks{
$^{1}$Benyamin Ghojogh and Mark Crowley are with the Machine Learning Laboratory, Department of Electrical and Computer Engineering, University of Waterloo, Waterloo, ON, Canada.
$^{2}$Ali Ghodsi is with the Department of Statistics and Actuarial Science, University of Waterloo, Waterloo, ON, Canada.
$^{3}$Fakhri Karray is with the Centre for Pattern Analysis and Machine Intelligence, Department of Electrical and Computer Engineering, University of Waterloo, Waterloo, ON, Canada.
Emails:
{\tt\small \{bghojogh, ali.ghodsi, karray, mcrowley\}@uwaterloo.ca
}
}%
}

\begin{document}

\maketitle
\thispagestyle{empty}
\pagestyle{empty}

\begin{abstract}

Locally Linear Embedding (LLE) is a nonlinear spectral dimensionality reduction and manifold learning method. It has two main steps which are linear reconstruction and linear embedding of points in the input space and embedding space, respectively. In this work, we propose two novel generative versions of LLE, named Generative LLE (GLLE), whose linear reconstruction steps are stochastic rather than deterministic. GLLE assumes that every data point is caused by its linear reconstruction weights as latent factors. The proposed GLLE algorithms can generate various LLE embeddings stochastically while all the generated embeddings relate to the original LLE embedding. We propose two versions for stochastic linear reconstruction, one using expectation maximization and another with direct sampling from a derived distribution by optimization. The proposed GLLE methods are closely related to and inspired by variational inference, factor analysis, and probabilistic principal component analysis. Our simulations show that the proposed GLLE methods work effectively in unfolding and generating submanifolds of data.  

\end{abstract}

\section{Introduction}

Dimensionality reduction and manifold learning methods are widely useful for feature extraction, manifold unfolding, and data visualization \cite{ghojogh2019feature}. The dimensionality reduction methods can be divided into three main categories, i.e., spectral methods, probabilistic methods, and neural network-based methods \cite{ghojogh2021data}. An example for spectral methods is Locally Linear Embedding (LLE) \cite{roweis2000nonlinear,saul2003think}. Examples for probabilistic methods are factor analysis \cite{fruchter1954introduction,child1990essentials} and probabilistic Principal Component Analysis (PCA) \cite{roweis1997algorithms,tipping1999probabilistic}. An example for neural network-based methods is variational autoencoder \cite{kingma2014auto} which formulates variational inference \cite{bishop2006pattern,blei2017variational,zhang2018advances,ghojogh2021factor} in an autoencoder framework. Another example for this category is adversarial autoencoder \cite{makhzani2015adversarial}. 

In this work, we propose two versions of Generative LLE (GLLE). GLLE is a combination of spectral and probabilistic methods. It is related to LLE, factor analysis, probabilistic PCA, and variational inference. Hence, we can say that it combines different categories of dimensionality reduction methods. The original LLE, which is a spectral method, has two main steps which are deterministic linear reconstruction of points in the input space and deterministic linear embedding in the embedding space. GLLE replaces deterministic linear reconstruction with a stochastic linear reconstruction while the linear embedding step is the same as in LLE. 

\begin{figure}[!t]
\centering
\includegraphics[width=0.5in]{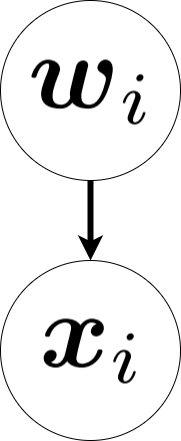}
\caption{The probabilistic graphical model for GLLE algorithms.}
\label{figure_PGM}
\end{figure}

In GLLE, we assume that reconstruction weights are latent factors causing the data points in a probabilistic graphical model; see Fig. \ref{figure_PGM}. Every data point is obtained by a function of its stochastic reconstruction weights. 
In the first proposed version of GLLE, the formulation of this function is similar to, but not the same as, the formulation in factor analysis and probabilistic PCA. The covariance matrices of the reconstruction weights are calculated using Expectation Maximization (ME) for the sake of Maximum Likelihood Estimation (MLE). In the second version of GLLE, we propose linear reconstruction with direct sampling in which the distribution of weights is obtained directly by some optimization derivations. The former approach is more solid in terms of theory while the latter is simpler to implement and slightly faster. 

The proposed GLLE algorithms can be categorized as generative models \cite{harshvardhan2020comprehensive}. It should be noted that they are not deep or autoencoder generative models \cite{oussidi2018deep,pan2019recent}. Rather, they are combinations of spectral and probabilistic approaches for dimensionality reduction. The proposed GLLE methods are useful for generating several LLE embeddings stochastically where the generated embeddings are related to the original LLE embedding. One can obtain more insights into the LLE embedding by generating more related embeddings and investigating into the new generated embeddings. Moreover, having a stochastic method rather than deterministic one gives the opportunity to have more number of embeddings for visualization and feature extraction without the need to several algorithms.

The remainder of this paper is organized as follows. Section \ref{section_background} reviews the related and required technical background. The proposed stochastic linear reconstruction with EM is explained in Section \ref{section_stochastic_linear_reconstruction_EM}. Section \ref{section_stochastic_linear_reconstruction_directSampling} explains the proposed stochastic linear reconstruction with direct sampling. The last step of GLLE, linear embedding, is explained in Section \ref{section_linear_embedding}. Experimental results are provided in Section \ref{section_experiments}. Finally, Section \ref{section_conclusion} concludes the paper. 

\section{Technical Background}\label{section_background}

In this section, we review some related technical background on joint distributions, linear reconstruction in LLE, variational inference, factor analysis, and probabilistic PCA.

\subsection{Marginal Multivariate Gaussian Distribution}

Consider two random variables $\b{x}_1 \in \mathbb{R}^{d_1}$ and $\b{x}_2 \in \mathbb{R}^{d_2}$ and let $\b{x}_3 := [\b{x}_1^\top, \b{x}_2^\top]^\top \in \mathbb{R}^{d_1 + d_2}$. Assume that $\b{x}_1$ and $\b{x}_2$ jointly multivariate Gaussian, i.e., $\b{x}_3 \sim \mathcal{N}(\b{x}_3; \b{\mu}_3, \b{\Sigma}_3)$.  The mean and covariance can be decomposed as:
\begin{align}
&\b{\mu}_3 =  [\b{\mu}_1^\top, \b{\mu}_2^\top]^\top \in \mathbb{R}^{d_1 + d_2}, \\
&\b{\Sigma}_3 = 
\begin{bmatrix}
\b{\Sigma}_{11} & \b{\Sigma}_{12} \\
\b{\Sigma}_{21} & \b{\Sigma}_{22}
\end{bmatrix}
\in \mathbb{R}^{(d_1 + d_2) \times (d_1 + d_2)}, \label{equation_factor_analysis_Sigma_y}
\end{align}
where $\b{\mu}_1 \in \mathbb{R}^{d_1}$, $\b{\mu}_2 \in \mathbb{R}^{d_2}$, $\b{\Sigma}_{11} \in \mathbb{R}^{d_1 \times d_2}$, $\b{\Sigma}_{22} \in \mathbb{R}^{d_2 \times d_2}$, $\b{\Sigma}_{12} \in \mathbb{R}^{d_1 \times d_2}$, and $\b{\Sigma}_{21} = \b{\Sigma}_{12}^\top$. 

It can be shown that the marginal distributions for $\b{x}_1$ and $\b{x}_2$ are Gaussian distributions where $\mathbb{E}[\b{x}_1] = \b{\mu}_1$ and $\mathbb{E}[\b{x}_2] = \b{\mu}_2$. 
The covariance matrix of the joint distribution can be simplified as \cite{ghojogh2021factor}:
\begin{align}\label{equation_covariance_marginal}
&\b{\Sigma}_3 \! =\! \mathbb{E}\Bigg[
\begin{bmatrix}
(\b{x}_1 - \b{\mu}_1)(\b{x}_1 - \b{\mu}_1)^\top, (\b{x}_1 - \b{\mu}_1)(\b{x}_2 - \b{\mu}_2)^\top \\
(\b{x}_2 - \b{\mu}_2)(\b{x}_1 - \b{\mu}_1)^\top, (\b{x}_2 - \b{\mu}_2)(\b{x}_2 - \b{\mu}_2)^\top
\end{bmatrix}
\Bigg],
\end{align}
where $\mathbb{E}[.]$ is the expectation operator. 
According to the definition of the multivariate Gaussian distribution, the conditional distribution is also a Gaussian distribution, i.e., $\b{x}_2 | \b{x}_1 \sim \mathcal{N}(\b{x}_2; \b{\mu}_{x_2|x_1}, \b{\Sigma}_{x_2|x_1})$ where \cite{ghojogh2021factor}:
\begin{align}
&\mathbb{R}^{d_2} \ni \b{\mu}_{x_2|x_1} := \b{\mu}_2 + \b{\Sigma}_{21}\b{\Sigma}_{11}^{-1} (\b{x}_1 - \b{\mu}_1), \label{equation_mean_conditional_in_joint} \\
&\mathbb{R}^{d_2 \times d_2} \ni \b{\Sigma}_{x_2|x_1} := \b{\Sigma}_{22} - \b{\Sigma}_{21}\b{\Sigma}_{11}^{-1} \b{\Sigma}_{12}, \label{equation_covariance_conditional_in_joint}
\end{align}
and likewise we have for $\b{x}_1 | \b{x}_2 \sim \mathcal{N}(\b{x}_1; \b{\mu}_{x_1|x_2}, \b{\Sigma}_{x_1|x_2})$.
Also, note that the probability density function of $d$-dimensional Gaussian distribution is:
\begin{align}\label{equation_multivariate_Gaussian_PDF}
&\mathcal{N}(\b{x}; \b{\mu}, \b{\Sigma})\! =\! \frac{1}{\sqrt{(2\pi)^d |\b{\Sigma}|}} \exp\Big(\!\!- \frac{(\b{x} - \b{\mu})^\top \b{\Sigma}^{-1} (\b{x} - \b{\mu})}{2}\Big),
\end{align}
where $|.|$ denotes the determinant of matrix. 

\subsection{Linear Reconstruction in Locally Linear Embedding}

A $k$NN graph is formed using pairwise Euclidean distance between the data points. Therefore, every data point has $k$ neighbors. Let $\b{x}_{ij} \in \mathbb{R}^d$ denote the $j$-th neighbor of $\b{x}_i  \in \mathbb{R}^d$ and let the matrix $\mathbb{R}^{d \times k} \ni \b{X}_i := [\b{x}_{i1}, \dots, \b{x}_{ik}]$ include the $k$ neighbors of $\b{x}_i$.
Suppose $\b{w}_i := [w_{i1}, \dots, w_{ik}]^\top \in \mathbb{R}^k$ denotes the reconstruction weights of every point $\b{x}_i$ by its neighbors $\b{X}_i$. 

The linear reconstruction in original LLE \cite{roweis2000nonlinear,saul2003think} is deterministic. 
In LLE, the weights for linear reconstruction of every point by its $k$NN are optimized as \cite{ghojogh2020locally}: 
\begin{equation}\label{equation_LLE_linearReconstruct}
\begin{aligned}
& \underset{\{\b{w}_i\}_{i=1}^n}{\text{minimize}}
& & \sum_{i=1}^n \Big|\Big|\b{x}_i - \sum_{j=1}^k w_{ij} \b{x}_{ij}\Big|\Big|_2^2, \\
& \text{subject to}
& & \sum_{j=1}^k w_{ij} = 1, ~~~ \forall i \in \{1, \dots, n\},
\end{aligned}
\end{equation}
which can be restated as:
\begin{equation}\label{equation_LLE_linearReconstruct_2}
\begin{aligned}
& \underset{\{\b{w}_i\}_{i=1}^n}{\text{minimize}}
& & \sum_{i=1}^n \b{w}_i^\top \b{G}_i\, \b{w}_i, \\
& \text{subject to}
& & \b{1}^\top \b{w}_i = 1, ~~~ \forall i \in \{1, \dots, n\},
\end{aligned}
\end{equation}
where $\mathbb{R}^{k \times k} \ni \b{G}_i := (\b{x}_i \b{1}^\top - \b{X}_i)^\top (\b{x}_i \b{1}^\top - \b{X}_i)$ and its solution is \cite{ghojogh2020locally}:
\begin{align}\label{equation_w_tilde_solution}
\b{w}_i = \frac{\b{G}_i^{-1} \b{1}}{\b{1}^\top \b{G}_i^{-1} \b{1}}.
\end{align}
The obtained weights $\{\b{w}_i\}_{i=1}^n$ are then used for linear embedding in the low dimensional space. In generative LLE, we replace the deterministic linear embedding with stochastic linear reconstruction.

\subsection{Variational Inference}

Assume data $\b{x}_i$ is caused by a latent factor $\b{w}_i$, as shown in Fig. \ref{figure_PGM}. Let the parameters of model be denoted by $\b{\theta}$.
In variational inference, the Evidence Lower Bound (ELBO) is a lower bound on the log likelihood of data and is defined as minus Kullback-Leibler (KL) divergence between a distribution $q(.)$ on $\b{w}_i$ and the joint distribution of $\b{x}_i$ and $\b{w}_i$. \cite{ghojogh2021factor,bishop2006pattern,kingma2014auto}:
\begin{align}\label{ELBO_equation1}
\mathcal{L}(q, \b{\theta}) := - \text{KL}\big(q(\b{w}_i)\, \|\, \mathbb{P}(\b{x}_i, \b{w}_i\, |\, \b{\theta})\big).
\end{align}
Maximizing this lower bound results in maximization of likelihood of data \cite{blei2017variational,zhang2018advances}. 
Variational inference uses EM for MLE \cite{ghojogh2021factor}:
\begin{align}
& q^{(t)}(\b{w}_i) \gets \mathbb{P}(\b{w}_i\, |\, \b{x}_i, \b{\theta}^{(t-1)}), \label{equation_E_step_variationalInference} \\
& \b{\theta}^{(t)} \gets \arg \max_{\b{\theta}}~ \mathbb{E}_{\sim q^{(t)}(\b{w}_i)} \big[\log \mathbb{P}(\b{x}_i, \b{w}_i\, |\, \b{\theta})\big], \label{equation_M_step_variationalInference}
\end{align}
where $\mathbb{E}[.]$ denotes expectation.
Here, we use the EM approach of variational inference for stochastic linear reconstruction in GLLE. 

\subsection{Factor Analysis}

Factor analysis \cite{fruchter1954introduction,child1990essentials} assumes that every data point $\b{x}_i$ is generated from a latent factor $\b{w}_i$. Its probabilistic graphical model is similar to Fig. \ref{figure_PGM} but with a small difference \cite{ghahramani1996algorithm}. 
It assumes $\b{x}_i$ is obtained by linear projection of $\b{w}_i$ onto the space of data by projection matrix $\b{\Lambda}$, then applying some linear translation, and finally adding a Gaussian noise $\b{\epsilon}$ with covariance matrix $\b{\Psi}$. This addition of noise is the main difference from the model depicted in Fig. \ref{figure_PGM}. If $\b{\mu}$ denotes the mean of data, factor analysis considers:
\begin{align}
&\b{x}_i := \b{\Lambda} \b{w}_i + \b{\mu} + \b{\epsilon}, \label{equation_factor_analysis} \\
&\mathbb{P}(\b{x}_i\, |\, \b{w}_i, \b{\Lambda}, \b{\mu}, \b{\Psi}) = \mathcal{N}(\b{x}_i; \b{\Lambda} \b{z}_i + \b{\mu}, \b{\Psi}), \label{equation_factor_analysis_likelihood_x_given_z}
\end{align}
where $\mathbb{P}(\b{w}_i) = \mathcal{N}(\b{w}_i; \b{0}, \b{I})$ and $\mathbb{P}(\b{\epsilon}) = \mathcal{N}(\b{\epsilon}; \b{0}, \b{\Psi})$. Factor analysis uses EM algorithm for finding optimum $\b{\Lambda}$ and $\b{\Psi}$ (see \cite{ghojogh2021factor} for details of EM in factor analysis).
In stochastic linear reconstruction for GLLE, we are inspired by Eq. (\ref{equation_factor_analysis}) for modeling the relation between data point and its reconstruction weights as its latent factor. 

\subsection{Probabilistic PCA}

Probabilistic PCA \cite{roweis1997algorithms,tipping1999probabilistic} is a special case of factor analysis where the variance of noise is equal in all dimensions of data space with covariance between dimensions, i.e.: 
\begin{align}\label{equation_PPCA_Psi}
\b{\Psi} = \sigma^2 \b{I}.
\end{align}
In other words, probabilistic PCA considers an isotropic noise model.
Similar to factor analysis, it can be solved iteratively using EM \cite{roweis1997algorithms}. However, one can also find a closed-form solution to its EM approach \cite{tipping1999probabilistic}.
Hence, by restricting the noise covariance to be isotropic, its solution becomes simpler and closed-form. See \cite{ghojogh2021factor} for details of derivations and closed-form solution for probabilistic PCA. 
Inspired by probabilistic PCA, in stochastic linear reconstruction for GLLE, we relax the covariance matrix of reconstruction weight to be diagonal. Interestingly, without this relaxation, the solution of M-step in EM algorithm of GLLE is solved iteratively; however, this relaxation makes the solution of M-step in EM algorithm of GLLE closed-form, as we also see the effect of relaxation in probabilistic PCA.

\section{Stochastic Linear Reconstruction with Expectation Maximization}\label{section_stochastic_linear_reconstruction_EM}

In this section, we propose the novel stochastic linear reconstruction for GLLE where EM is used for MLE and calculating the covariance matrices of the reconstruction weights of points. 

\subsection{Notations and Joint and Conditional Distributions}

As Fig. \ref{figure_PGM} depicts, we assume that the every point $\b{x}_i$ is generated by its reconstruction weights $\b{w}_i$ as a latent factor. 
Therefore, $\b{x}_i$ can be written as a stochastic function of $\b{w}_i$ where we assume $\b{w}_i$ has a multivariate Gaussian distribution:
\begin{align}
&\b{x}_i = \b{X}_i \b{w}_i + \b{\mu}, \label{equation_x_X_w} \\
&\mathbb{P}(\b{w}_i) = \mathcal{N}(\b{w}_i; \b{0}, \b{\Omega}_i) \implies \mathbb{E}[\b{w}_i] = \b{0}, ~~\mathbb{E}[\b{w}_i \b{w}_i^\top] = \b{\Omega}_i, \label{equation_prior_w} 
\end{align}
where $\b{\Omega}_i \in \mathbb{R}^{k \times k}$ is covariance of $\b{w}_i$ and $\b{\mu} \in \mathbb{R}^d$ is the mean of data because:
\begin{align}
\frac{1}{n} \sum_{i=1}^n \b{x}_i = \mathbb{E}[\b{x}_i] = \b{X}_i \mathbb{E}[\b{w}_i] + \b{\mu} \overset{(\ref{equation_prior_w})}{=} \b{0} + \b{\mu} = \b{\mu}.
\end{align}
According to Eq. (\ref{equation_covariance_marginal}), for the joint distribution of $[\b{x}_i^\top, \b{w}_i^\top]^\top \in \mathbb{R}^{d + k}$,  we have:
\begin{alignat*}{2}
&\b{\Sigma}_{11} &&= \mathbb{E}[(\b{x}_i - \b{\mu}) (\b{x}_i - \b{\mu})^\top] \overset{(\ref{equation_x_X_w})}{=} \mathbb{E}[(\b{X}_i \b{w}_i) (\b{X}_i \b{w}_i)^\top] \\
& &&= \b{X}_i \mathbb{E}[\b{w}_i \b{w}_i^\top] \b{X}_i^\top \overset{(\ref{equation_prior_w})}{=} \b{X}_i \b{\Omega}_i \b{X}_i^\top, \\
&\b{\Sigma}_{12} &&= \mathbb{E}[(\b{x}_i - \b{\mu}) (\b{w}_i - \b{0})^\top] \overset{(\ref{equation_x_X_w})}{=} \b{X}_i \mathbb{E}[\b{w}_i \b{w}_i^\top] \overset{(\ref{equation_prior_w})}{=} \b{X}_i \b{\Omega}_i, \\
&\b{\Sigma}_{22} &&= \mathbb{E}[(\b{w}_i - \b{0}) (\b{w}_i - \b{0})^\top] = \mathbb{E}[\b{w}_i \b{w}_i^\top] \overset{(\ref{equation_prior_w})}{=} \b{\Omega}_i.
\end{alignat*}
Hence:
\begin{align}\label{equation_joint_distribution_x_w}
\begin{bmatrix}
\b{x}_{i} \\
\b{w}_{i}
\end{bmatrix}
\sim
\mathcal{N}\Bigg(
\begin{bmatrix}
\b{x}_i \\
\b{w}_i 
\end{bmatrix}
;
\begin{bmatrix}
\b{\mu} \\
\b{0} 
\end{bmatrix}
,
\begin{bmatrix}
\b{X}_i \b{\Omega}_i \b{X}_i^\top & \b{X}_i \b{\Omega}_i \\
\b{\Omega}_i^\top \b{X}_i^\top & \b{\Omega}_i
\end{bmatrix}
\Bigg).
\end{align}
We have:
\begin{align}\label{equation_x_given_w}
&\mathbb{P}(\b{x}_i\, |\, \b{w}_i, \b{\Omega}_i) \overset{(a)}{=} \mathcal{N}(\b{x}_i; \b{X}_i \b{w}_i + \b{\mu}, \b{X}_i \b{\Omega}_i \b{X}_i^\top),
\end{align}
where $(a)$ is because of Eqs. (\ref{equation_x_X_w}) and (\ref{equation_joint_distribution_x_w}).
We use EM for MLE in stochastic linear reconstruction. In the following, the steps of EM are explained.  

\subsection{E-Step in Expectation Maximization}

As we will see later in the M-step of EM, we will have two expectation terms which need to be computed in the E-step. These expectations, which are over the $q(\b{w}_i) := \mathbb{P}(\b{w}_i\, |\, \b{x}_i)$ distribution, are $\mathbb{E}_{\sim q^{(t)}(\b{w}_i)}[\b{w}_i] \in \mathbb{R}^k$ and $\mathbb{E}_{\sim q^{(t)}(\b{w}_i)}[\b{w}_i \b{w}_i^\top] \in \mathbb{R}^{k \times k}$ where $t$ denotes the iteration index in EM iterations.
According to Eqs. (\ref{equation_mean_conditional_in_joint}), (\ref{equation_covariance_conditional_in_joint}), and (\ref{equation_joint_distribution_x_w}), we have:
\begin{align}
&\mathbb{E}_{\sim q^{(t)}(\b{w}_i)}[\b{w}_i] = \b{\mu}_{w|x} = \b{\Omega}_i^\top \b{X}_i^\top (\b{X}_i \b{\Omega}_i \b{X}_i^\top)^{\dagger} (\b{x}_i - \b{\mu}), \label{E_step_expectation_w} \\
&\mathbb{E}_{\sim q^{(t)}(\b{w}_i)}[\b{w}_i \b{w}_i^\top] = \b{\Sigma}_{w|x} \nonumber \\
&~~~~~~~~~~~~~~~ = \b{\Omega}_i - \b{\Omega}_i^\top \b{X}_i^\top (\b{X}_i \b{\Omega}_i \b{X}_i^\top)^{\dagger} \b{X}_i \b{\Omega}_i, \label{E_step_expectation_w_w} 
\end{align}
where $^\dagger$ denotes either inverse or pseudo-inverse of matrix. 

\subsection{M-Step in Expectation Maximization}

In M-step of EM, we maximize the joint likelihood of data and weights over all $n$ data points where the optimization variable is the covariance of prior distribution of weights:
\begin{align*}
& \max_{\{\b{\Omega}_i\}_{i=1}^n}~ \sum_{i=1}^n \mathbb{E}_{\sim q^{(t)}(\b{w}_i)} \big[\log \mathbb{P}(\b{x}_i, \b{w}_i\, |\, \b{\Omega}_i)\big] \nonumber \\
&~~~~ \overset{(a)}{=} \max_{\{\b{\Omega}_i\}_{i=1}^n}~ \sum_{i=1}^n \Big( \mathbb{E}_{\sim q^{(t)}(\b{w}_i)} \big[\log \mathbb{P}(\b{x}_i\, |\, \b{w}_i, \b{\Omega}_i)\big] \nonumber \\
&~~~~~~~~~~~~~~~~~~~~~~~~~~~~ + \mathbb{E}_{\sim q^{(t)}(\b{w}_i)} \big[\log \mathbb{P}(\b{w}_i)\big] \Big) \\
& \overset{(b)}{=} \max_{\{\b{\Omega}_i\}_{i=1}^n} \sum_{i=1}^n \Big( \mathbb{E}_{\sim q^{(t)}(\b{w}_i)} \big[\log \mathcal{N}(\b{X}_i \b{w}_i + \b{\mu}, \b{X}_i \b{\Omega}_i \b{X}_i^\top) \big]  \\
&~~~~~~~~~~~~~~~~~~~~~~~~~~~~ + \mathbb{E}_{\sim q^{(t)}(\b{w}_i)} \big[\log \mathcal{N}(\b{0}, \b{\Omega}_i) )\big] \Big) 
\end{align*}
\begin{align*}
&\overset{(\ref{equation_multivariate_Gaussian_PDF})}{=} \max_{\{\b{\Omega}_i\}_{i=1}^n}~ \Big( \underbrace{-\frac{d\,n}{2} \log(2\pi)}_\text{constant} - \frac{n}{2} \log |\b{X}_i \b{\Omega}_i \b{X}_i^\top|  \\
&~~~~ - \frac{1}{2} \sum_{i=1}^n \mathbb{E}_{\sim q^{(t)}(\b{w}_i)} \big[ (\b{x}_i - \b{X}_i\b{w}_i - \b{\mu})^\top (\b{X}_i \b{\Omega}_i \b{X}_i^\top)^{-1} \\
&~~~~~~~~~~~~~~~~ (\b{x}_i - \b{X}_i\b{w}_i - \b{\mu}) \big] -\underbrace{\frac{k\,n}{2} \log(2\pi)}_\text{constant} \\ 
&~~~~ - \frac{n}{2} \log |\b{\Omega}_i| - \frac{1}{2} \sum_{i=1}^n \mathbb{E}_{\sim q^{(t)}(\b{w}_i)} \big[ \b{w}_i^\top \b{\Omega}_i^{-1} \b{w}_i \big] \Big) 
\end{align*}
\begin{align}
&\overset{(c)}{=} \max_{\{\b{\Omega}_i\}_{i=1}^n} \Big(\!\! - \frac{n}{2} \log |\b{X}_i \b{\Omega}_i \b{X}_i^\top| - \frac{n}{2} \textbf{tr}\big((\b{X}_i \b{\Omega}_i \b{X}_i^\top)^{-1} \b{S}_1\big) \nonumber \\
&~~~~~~~~~~~~~~ - \frac{n}{2} \log |\b{\Omega}_i| - \frac{n}{2} \textbf{tr}\big( \b{\Omega}_i^{-1} \b{S}_2 \big) \Big), \label{equation_joint_likelihood}
\end{align}
where $(a)$ is because of the chain rule $\mathbb{P}(\b{x}_i, \b{w}_i\, |\, \b{\Omega}_i) = \mathbb{P}(\b{x}_i\, |\, \b{w}_i, \b{\Omega}_i)\, \mathbb{P}(\b{w}_i)$, and $(b)$ is because of Eqs. (\ref{equation_prior_w}) and (\ref{equation_x_given_w}), and $(c)$ is because we define the scatters $\b{S}_1$ and $\b{S}_2$ as:
\begin{align}
&\mathbb{R}^{d \times d} \ni \b{S}_1 := \frac{1}{n} \sum_{i=1}^n \mathbb{E}_{\sim q^{(t)}(\b{w}_i)} \big[ (\b{x}_i - \b{X}_i\b{w}_i - \b{\mu}) \nonumber \\
&~~~~ (\b{x}_i - \b{X}_i\b{w}_i - \b{\mu})^\top \big] = \frac{1}{n} \sum_{i=1}^n \Big( (\b{x}_i - \b{\mu}) (\b{x}_i - \b{\mu})^\top \nonumber \\
&~~~~~~~~~~ - 2 \b{X}_i \mathbb{E}_{\sim q^{(t)}(\b{w}_i)} \big[ \b{w}_i \big] (\b{x}_i - \b{\mu})^\top \nonumber \\
&~~~~~~~~~~ + \b{X}_i \mathbb{E}_{\sim q^{(t)}(\b{w}_i)} \big[ \b{w}_i \b{w}_i^\top \big] \b{X}_i^\top \Big), \label{equation_S1}\\
&\mathbb{R}^{k \times k} \ni \b{S}_2 := \frac{1}{n} \sum_{i=1}^n \mathbb{E}_{\sim q^{(t)}(\b{w}_i)} \big[ \b{w}_i \b{w}_i^\top \big], \label{equation_S2}
\end{align}
where the expectation terms are found by Eqs. (\ref{E_step_expectation_w}) and (\ref{E_step_expectation_w_w}). 
The gradient of the joint likelihood is:
\begin{align}
\mathbb{R}^{k \times k} \ni &\frac{\partial \text{ Eq. } (\ref{equation_joint_likelihood})}{\partial \b{\Omega}_i^{-1}} = \frac{n}{2} \Big[ \textbf{vec}^{-1}_{k \times k} \big[ \b{T}_i\, \textbf{vec}_{d^2 \times 1}(\b{X}_i \b{\Omega}_i \b{X}_i^\top) \big] \nonumber \\
&- \textbf{vec}^{-1}_{k \times k} \big[ \b{T}_i\, \textbf{vec}_{d^2 \times 1}(\b{S}_1) \big] + \b{\Omega}_i - \b{S}_2 \Big], \label{equation_gradient_joint_likelihood}
\end{align}
where we use the Magnus-Neudecker convention in which matrices are vectorized, $\textbf{vec}(.)$ vectorizes the matrix, $\textbf{vec}^{-1}_{k \times k}(.)$ is de-vectorization to $k \times k$ matrix, $\otimes$ denotes the Kronecker product, and $\mathbb{R}^{k^2 \times d^2} \ni \b{T}_i := \b{X}_i^\top \otimes \b{X}_i^\top$.

In the M-step, one can update the variables $\{\b{\Omega}_i\}_{i=1}^n$ using gradient descent \cite{nocedal2006numerical} with the gradient in Eq. (\ref{equation_gradient_joint_likelihood}). However, inspired by probabilistic PCA, we can relax the covariance matrix and simplify the algorithm. 

\subsection{Relaxation of Covariance}

Inspired by relaxation of factor analysis for probabilistic PCA, we can relax the covariance matrix to be diagonal and the variance of weights to be equal in all $k$ dimensions: 
\begin{align}\label{equation_diagonal_Omega_covariance}
\b{\Omega}_i = \sigma_i \b{I} \in \mathbb{R}^{k \times k}.
\end{align}
Substituting this covariance into Eq. (\ref{equation_joint_likelihood}) and noticing the properties of determinant and trace gives:
\begin{align}
&\max_{\{\sigma_i\}_{i=1}^n} \Big(\!\! - \frac{n}{2} \log (\sigma_i^d |\b{X}_i \b{X}_i^\top|) - \frac{n}{2} \sigma_i^{-1} \textbf{tr}\big((\b{X}_i \b{X}_i^\top)^{-1} \b{S}_1\big) \nonumber \\
&~~~~~~~~~~~~~~ - \frac{n}{2} \log \sigma_i^k - \frac{n}{2} \sigma_i^{-1} \textbf{tr}\big( \b{S}_2 \big) \Big) \nonumber \\
&\overset{(a)}{=} \max_{\{\sigma_i\}_{i=1}^n} \Big(\!\! - \frac{n}{2} (d + k) \log (\sigma_i) \nonumber \\
&~~~~~~~~~~ - \frac{n}{2} \Big[ \textbf{tr}\big((\b{X}_i \b{X}_i^\top)^{-1} \b{S}_1\big) - \textbf{tr}\big( \b{S}_2 \big) \Big] \sigma_i^{-1} \Big), \label{equation_joint_likelihood_relaxed}
\end{align}
where $(a)$ is because $\log (\sigma_i^d |\b{X}_i \b{X}_i^\top|) = d\log (\sigma_i) + \log (|\b{X}_i \b{X}_i^\top|)$ whose second term is a constant w.r.t. $\sigma_i$. 
Setting the gradient of the joint likelihood to zero gives:
\begin{align}
&\mathbb{R} \ni \frac{\partial \text{ Eq. } (\ref{equation_joint_likelihood_relaxed})}{\partial \sigma_i^{-1}} = \frac{n}{2} \Big[ (d + k) \sigma_i \nonumber \\
&~~~~~~~~~~ - \Big( \textbf{tr}\big((\b{X}_i \b{X}_i^\top)^{-1} \b{S}_1\big) + \textbf{tr}(\b{S}_2) \Big) \Big] \overset{\text{set}}{=} 0 \nonumber \\
&\implies \sigma_i = (d + k)^{-1} \Big( \textbf{tr}\big((\b{X}_i \b{X}_i^\top)^{\dagger} \b{S}_1\big) + \textbf{tr}(\b{S}_2) \Big), \label{equation_variance_M_step}
\end{align}
where $^\dagger$ denotes either inverse or pseudo-inverse of matrix. 
The EM algorithm for stochastic linear reconstruction in GLLE is summarized in Algorithm \ref{algorithm_linear_reconstruction_EM}.
We can sample $\{\b{w}_i\}_{i=1}^n$ with the following prior and conditional distributions:
\begin{align}
& \b{w}_i \sim \mathcal{N}(\b{w}_i; \b{0}, \sigma_i \b{I}), \\
& \b{w}_i\, |\, \b{x}_i \sim \mathcal{N}(\b{w}_i; \b{\mu}_{w | x}, \b{\Sigma}_{w | x}), \label{equation_distribution_w_given_x}
\end{align}
where $\b{\mu}_{w | x}$ and $\b{\Sigma}_{w | x}$ are defined in Eqs. (\ref{E_step_expectation_w}) and (\ref{E_step_expectation_w_w}), respectively. 


\SetAlCapSkip{0.5em}
\IncMargin{0.8em}
\begin{algorithm2e}[!t]
\DontPrintSemicolon
    \textbf{Input: } $k$NN graph or $\{\b{X}_i\}_{i=1}^n$, $\{\b{x}_i\}_{i=1}^n$ \;
    \textbf{Initialize: } $\{\b{\Omega}_i\}_{i=1}^n = \b{I}$\;
    \While{not converged}{
        // E-step:\;
        \For{every data point $i$ from $1$ to $n$}{
            Calculate expectations by Eqs. (\ref{E_step_expectation_w}) and (\ref{E_step_expectation_w_w})\;
        }
        // Sampling:\;
        Sample weights $\{\b{w}_i\}_{i=1}^n$ using Eq. (\ref{equation_distribution_w_given_x})\;
        // M-step:\;
        Calculate $\b{S}_1$ and $\b{S}_2$ using Eqs. (\ref{equation_S1}) and (\ref{equation_S2})\;
        \For{every data point $i$ from $1$ to $n$}{
            Calculate $\sigma_i$ by Eq. (\ref{equation_variance_M_step})\;
            Calculate $\b{\Omega}_i$ using Eq. (\ref{equation_diagonal_Omega_covariance})\;
        }
    }
    \textbf{Return} weights $\{\b{w}_i\}_{i=1}^n$\;
\caption{Stochastic Linear Reconstruction with Expectation Maximization}\label{algorithm_linear_reconstruction_EM}
\end{algorithm2e}
\DecMargin{0.8em}

\section{Stochastic Linear Reconstruction with Direct Sampling}\label{section_stochastic_linear_reconstruction_directSampling}

In this section, we propose the novel stochastic linear reconstruction for GLLE where the distributions of the reconstruction weights of points are calculated by derivations of optimization and the weights are sampled directly from those distributions. 

The stochastic linear reconstruction with direct sampling is initialized by the original LLE \cite{roweis2000nonlinear,saul2003think}. 
Assume we have the embedding of LLE; therefore, we have $\{\b{X}_i \in \mathbb{R}^{d \times k}\}_{i=1}^n$ and $\{\b{Y}_i \in \mathbb{R}^{p \times k}\}_{i=1}^n$ from the $k$NN graph of data and the LLE embedding, respectively. Note that columns of $\b{X}_i$ are the neighbor points to $\b{x}_i$ and columns of $\b{Y}_i$ are the embeddings of $\b{X}_i$. From the original LLE, we also have the embeddings and reconstruction weights of points, denoted by $\{\b{y}_i\}_{i=1}^n$ and $\{\b{w}_i^\text{LLE}\}_{i=1}^n$, respectively. 

In stochastic linear reconstruction with direct sampling, we find the reconstruction weights in a way that every point is reconstructed best by a linear combination of its neighbors in both input and embedded spaces:
\begin{align}\label{equation_stochastic_linear_reconstruction_directSampling_opt1}
\underset{\{\b{w}_i\}_{i=1}^n}{\text{minimize}} \quad \sum_{i=1}^n \Big( \|\b{x}_i - \b{X}_i \b{w}_i\|_2^2 + \|\b{y}_i - \b{Y}_i \b{w}_i\|_2^2 \Big).
\end{align}
This problem can be seen as minimization for every point. 
We assume a $k$-dimensional Gaussian distribution for the conditional distribution $\mathbb{P}(\b{w}_i\, |\, \b{y}_i, \b{x}_i)$. 
The minimization (\ref{equation_stochastic_linear_reconstruction_directSampling_opt1}) can be restated to the following maximization:
\begin{align}
&\underset{\b{w}_i}{\text{maximize}} \quad \mathbb{P}(\b{w}_i\, |\, \b{y}_i, \b{x}_i) \propto \nonumber \\
&~~~~\exp\Big(\!\!-\! \big( \|\b{x}_i - \b{X}_i \b{w}_i\|_2^2 + \|\b{y}_i - \b{Y}_i \b{w}_i\|_2^2 \big)\Big).
\end{align}
We can simplify the term in the exponential as:
\begin{align}
\|&\b{x}_i - \b{X}_i \b{w}_i\|_2^2 + \|\b{y}_i - \b{Y}_i \b{w}_i\|_2^2 = \b{x}_i^\top \b{x}_i - 2\b{x}_i^\top \b{X}_i \b{w}_i \nonumber \\
&+ \b{w}_i^\top \b{X}_i^\top \b{X}_i \b{w}_i + \b{y}_i^\top \b{y}_i - 2\b{y}_i^\top \b{Y}_i \b{w}_i + \b{w}_i^\top \b{Y}_i^\top \b{Y}_i \b{w}_i \nonumber \\
&= \b{x}_i^\top \b{x}_i + \b{y}_i^\top \b{y}_i + \b{w}_i^\top (\b{X}_i^\top \b{X}_i + \b{Y}_i^\top \b{Y}_i) \b{w}_i \nonumber \\
& - 2 \b{y}_i^\top \b{Y}_i \b{w}_i - 2 \b{x}_i^\top \b{X}_i \b{w}_i \nonumber \\
&\propto \big(\b{w}_i - (\b{X}_i^\top \b{X}_i + \b{Y}_i^\top \b{Y}_i)^{-1} (\b{X}_i^\top \b{x}_i + \b{Y}_i^\top \b{y}_i) \big)^\top \nonumber \\
&~~~~(\b{X}_i^\top \b{X}_i + \b{Y}_i^\top \b{Y}_i) \nonumber \\
&~~~~\big(\b{w}_i - (\b{X}_i^\top \b{X}_i + \b{Y}_i^\top \b{Y}_i)^{-1} (\b{X}_i^\top \b{x}_i + \b{Y}_i^\top \b{y}_i)\big).
\end{align}
Therefore, according to Eq. (\ref{equation_multivariate_Gaussian_PDF}):
\begin{align}
&\mathbb{R}^{k \times k} \ni \b{\Gamma}_i := \text{Cov}(\b{w}_i\, |\, \b{y}_i, \b{x}_i) = (\b{X}_i^\top \b{X}_i + \b{Y}_i^\top \b{Y}_i)^{-1}, \nonumber \\
& \mathbb{R}^{k} \ni \mathbb{E}[\b{w}_i\, |\, \b{y}_i, \b{x}_i] = \b{\Gamma}_i (\b{X}_i^\top \b{x}_i + \b{Y}_i^\top \b{y}_i), \nonumber \\
& \mathbb{P}(\b{w}_i\, |\, \b{y}_i, \b{x}_i) = \mathcal{N}(\b{w}_i; \mathbb{E}[\b{w}_i\, |\, \b{y}_i, \b{x}_i], \b{\Gamma}_i). \label{equation_stochastic_linear_reconstruction_directSampling_w_given_y_and_x}
\end{align}
Note that, in stochastic linear reconstruction with direct sampling, we already have the weights obtained from the original LLE. 
Hence, we have the true mean of distribution of $\b{w}_i$ which is the weight obtained from the original LLE, denoted by $\b{w}_i^\text{LLE}$. 
Therefore, we can relax this distribution as:
\begin{align}
\mathbb{P}(\b{w}_i\, |\, \b{y}_i, \b{x}_i) = \mathcal{N}(\b{w}_i; \b{w}_i^\text{LLE}, \b{\Gamma}_i). \label{equation_stochastic_linear_reconstruction_directSampling_w_given_y_and_x_relaxed}
\end{align}
The algorithm of stochastic linear reconstruction with direct sampling is Algorithm \ref{algorithm_linear_reconstruction_directSampling}. Note that this direct sampling is less theoretically solid than the EM approach for stochastic linear reconstruction; however, it is simpler to implement and slightly faster.

\SetAlCapSkip{0.5em}
\IncMargin{0.8em}
\begin{algorithm2e}[!t]
\DontPrintSemicolon
    \textbf{Input: } dataset $\{\b{x}_i\}_{i=1}^n$ \;
    Find $\{\b{y}_{i}\}_{i=1}^n$ and $\{\b{w}_{i}^\text{LLE}\}_{i=1}^n$ using original LLE\;
    \For{instance $i$ from $1$ to $n$}{
        Sample $\b{w}_{i} \sim \mathbb{P}(\b{w}_{i}\, |\, \b{y}_{i}, \b{x}_i)$ using Eq. (\ref{equation_stochastic_linear_reconstruction_directSampling_w_given_y_and_x_relaxed})\;
    }
    \textbf{Return} weights $\{\b{w}_i\}_{i=1}^n$\;
\caption{Stochastic Linear Reconstruction with Direct Sampling}\label{algorithm_linear_reconstruction_directSampling}
\end{algorithm2e}
\DecMargin{0.8em}

\section{Linear Embedding}\label{section_linear_embedding}

Using either EM or direct sampling, we found the weights, $\{\b{w}_i := [w_{i1}, \dots, w_{ik}]^\top\}_{i=1}^n$, stochastically for linear reconstruction in the high dimensional input space. 
Now we embed data in the low dimensional embedding space using the same weights as in the input space. 
This linear embedding, which is the same as linear embedding in deterministic LLE \cite{roweis2000nonlinear,saul2003think}, can be formulated as the following optimization \cite{ghojogh2020locally}: 
\begin{equation}\label{equation_LLE_linearEmbedding}
\begin{aligned}
& \underset{\b{Y}}{\text{minimize}}
& & \sum_{i=1}^n \Big|\Big|\b{y}_i - \sum_{j=1}^n \breve{w}_{ij} \b{y}_j\Big|\Big|_2^2, \\
& \text{subject to}
& & \frac{1}{n} \sum_{i=1}^n \b{y}_i \b{y}_i^\top = \b{I}, \quad \sum_{i=1}^n \b{y}_i = \b{0}, 
\end{aligned}
\end{equation}
where the rows of $\mathbb{R}^{n \times p} \ni \b{Y} := [\b{y}_1, \dots, \b{y}_n]^\top$ are the embedded data points (stacked row-wise), $\b{y}_i \in \mathbb{R}^p$ is the $i$-th embedded data point, and $\breve{w}_{ij}$ is the weight obtained from the stochastic linear reconstruction if $\b{x}_j$ is a neighbor of $\b{x}_i$ and zero otherwise:
\begin{align}\label{equaion_LLE_weight_and_weightHat}
\breve{w}_{ij} := 
\left\{
    \begin{array}{ll}
        w_{ij} & \mbox{if } \b{x}_j \in k\text{NN}(\b{x}_i), \\
        0 & \mbox{otherwise}.
    \end{array}
\right.
\end{align}
The second constraint in Eq. (\ref{equation_LLE_linearEmbedding}) ensures the zero mean of embedded data points. The first and second constraints together satisfy having unit covariance for the embedded points.

\begin{figure}[!t]
\centering
\includegraphics[width=2.5in]{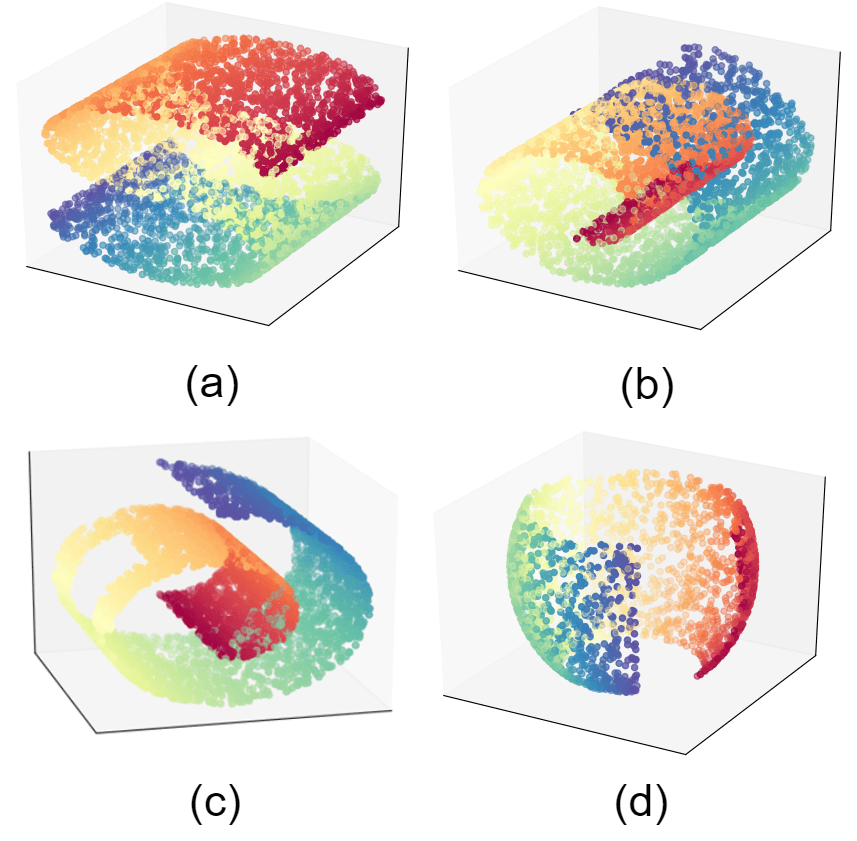}
\caption{Three synthetic nonlinear manifolds used for unfolding experiments: (a) S-curve, (b) Swiss roll, (c) Swiss roll with hole, and (d) severed bowl.}
\label{figure_datasets}
\end{figure}

\begin{figure*}[!t]
\centering
\includegraphics[width=\textwidth]{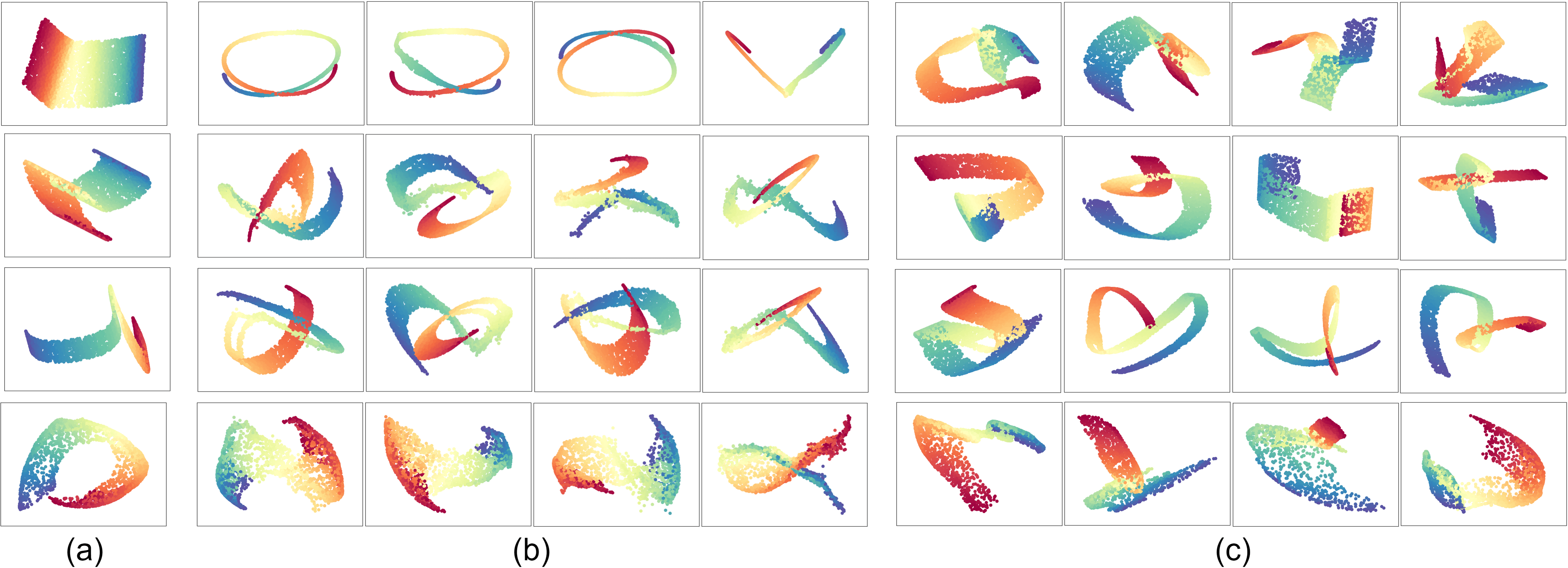}
\caption{(a) LLE unfolding and several unfolding generations of GLLE embedding with (b) EM algorithm and (c) direct sampling for the three nonlinear datasets. The first to fourth rows correspond to the S-curve, Swiss roll, Swiss roll with hole, and severed bowl, respectively.}
\label{figure_GLLE_generations}
\end{figure*}

\begin{figure*}[!t]
\centering
\includegraphics[width=\textwidth]{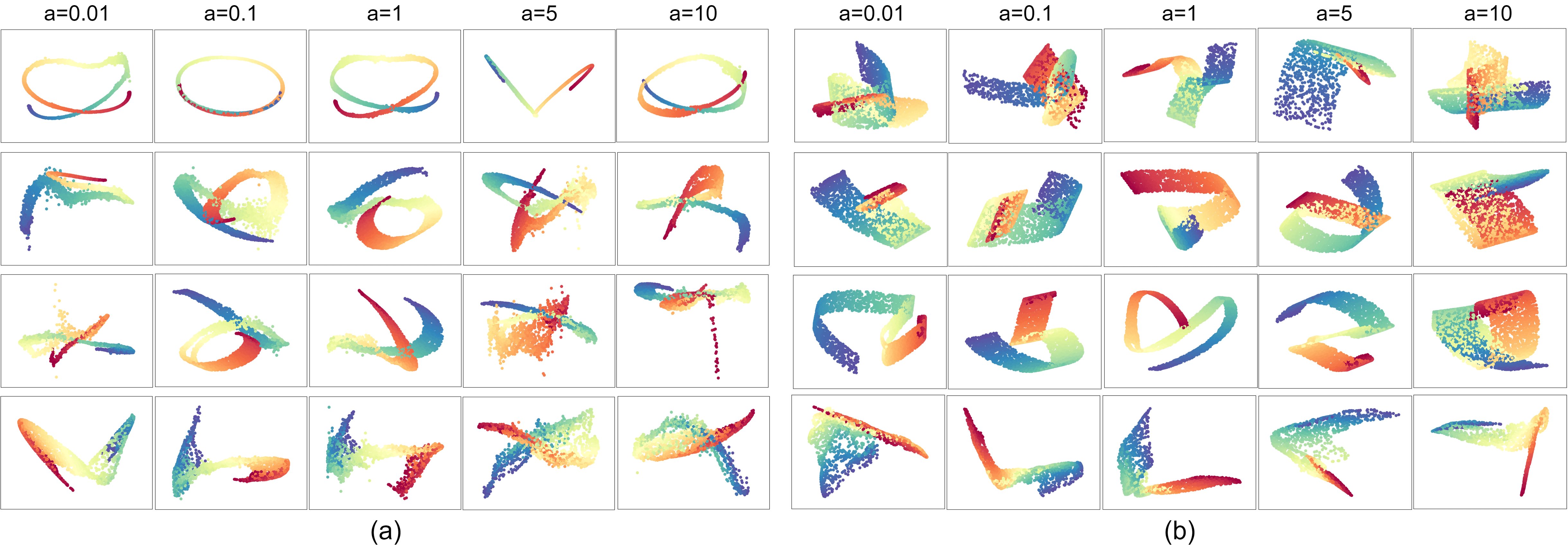}
\caption{The effect of covariance scaling on manifold unfolding by (a) GLLE with EM algorithm and (b) GLLE with direct sampling. The first to fourth rows correspond to S-curve, Swiss roll, Swiss roll with hole, and severed bowl, respectively. The columns correspond to different covariance scales, which are (a) $a \b{\Omega}_i$ and (b) $a \b{\Gamma}_i$ for $\forall i$ and $a \in \{0.01, 0.1, 1, 5, 10\}$.}
\label{figure_GLLE_interpolation}
\end{figure*}

Suppose $\mathbb{R}^n \ni \breve{\b{w}}_i := [\breve{w}_{i1}, \dots, \breve{w}_{in}]^\top$ and let $\mathbb{R}^n \ni \b{1}_i := [0, \dots, 1, \dots, 0]^\top$ be the vector whose $i$-th element is one and other elements are zero.
The objective function in Eq. (\ref{equation_LLE_linearEmbedding}) can be restated as:
\begin{align*}
&\sum_{i=1}^n \Big|\Big|\b{y}_i - \sum_{j=1}^n \breve{w}_{ij} \b{y}_j\Big|\Big|_2^2 = \sum_{i=1}^n ||\b{Y}^\top\b{1}_i - \b{Y}^\top\breve{\b{w}}_i||_2^2 \\
&= ||\b{Y}^\top\b{I} - \b{Y}^\top\breve{\b{W}}^\top||_F^2 = ||\b{Y}^\top (\b{I} - \breve{\b{W}})^\top||_F^2 \\
&= ||\b{Y}^\top(\b{I} - \breve{\b{W}})^\top||_F^2 = \textbf{tr}\big((\b{I} - \breve{\b{W}})\b{Y}\b{Y}^\top(\b{I} - \breve{\b{W}})^\top\big) \\
&= \textbf{tr}\big(\b{Y}^\top(\b{I} - \breve{\b{W}})^\top(\b{I} - \breve{\b{W}})\b{Y}\big) = \textbf{tr}(\b{Y}^\top\b{M}\b{Y}),
\end{align*}
where $||.||_F$ denotes the Frobenius norm of matrix, the $i$-th row of $\mathbb{R}^{n \times n} \ni \breve{\b{W}} := [\breve{\b{w}}_1, \dots, \breve{\b{w}}_n]^\top$ includes the weights for the $i$-th data point and:
\begin{align}\label{equation_M}
\mathbb{R}^{n \times n} \ni \b{M} := (\b{I} - \breve{\b{W}})^\top (\b{I} - \breve{\b{W}}).
\end{align}
Finally, Eq. (\ref{equation_LLE_linearEmbedding}) can be rewritten as:
\begin{equation}\label{equation_LLE_linearEmbedding_2}
\begin{aligned}
& \underset{\b{Y}}{\text{minimize}}
& & \textbf{tr}(\b{Y}^\top\b{M}\b{Y}), \\
& \text{subject to}
& & \frac{1}{n} \b{Y}^\top \b{Y} = \b{I}, \quad \b{Y}^\top \b{1} = \b{0}, 
\end{aligned}
\end{equation}
The second constraint is satisfied implicitly \cite{ghojogh2020locally}. The Lagrangian for Eq. (\ref{equation_LLE_linearEmbedding_2}) is \cite{boyd2004convex}:
\begin{align*}
\mathcal{L} = \textbf{tr}(\b{Y}^\top\b{M}\b{Y}) - \textbf{tr}\big(\b{\Lambda}^\top (\frac{1}{n} \b{Y}^\top \b{Y} - \b{I})\big),
\end{align*}
where $\b{\Lambda} \in \mathbb{R}^{n \times n}$ is a diagonal matrix including the Lagrange multipliers. 
Equating derivative of $\mathcal{L}$ to zero gives us:
\begin{align}
&\mathbb{R}^{n \times p} \ni \frac{\partial \mathcal{L}}{\partial \b{Y}} = 2\b{M}\b{Y} - \frac{2}{n} \b{Y} \b{\Lambda} \overset{\text{set}}{=} \b{0} \nonumber \\
&\implies \b{M}\b{Y} = \b{Y} (\frac{1}{n}\b{\Lambda}), \label{equation_LLE_linearEmbedding_eigenproblem}
\end{align}
which is the eigenvalue problem for $\b{M}$ \cite{ghojogh2019eigenvalue}. Therefore, the columns of $\b{Y}$ are the eigenvectors of $\b{M}$ where eigenvalues are the diagonal elements of $(1/n)\b{\Lambda}$.
As Eq. (\ref{equation_LLE_linearEmbedding_2}) is a minimization problem, the columns of $\b{Y}$ should be sorted from the smallest to largest corresponding non-zero eigenvalues. Note that because of relation of $\b{M}$ with Laplacian of $k$NN graph (see Eq. (\ref{equation_M})), there is a zero eigenvalue whose eigenvector should be ignored.

\section{Simulations}\label{section_experiments}

In this section, we report the simulation results.
The code for this paper and its simulations can be found in our Github repository\footnote{\texttt{https://github.com/bghojogh/Generative-LLE}}.

\subsection{Datasets}

For evaluating the effectiveness of the proposed GLLE with EM algorithm and direct sampling and its comparison with original LLE embedding, we created four highly nonlinear manifolds. The created synthetic datasets are S-curve, Swiss roll, Swiss roll with a hole inside it, and severed bowl; they are illustrated in Fig. \ref{figure_datasets}. Each of these nonlinear datasets includes 5000 three-dimensional data points.

\subsection{Manifold Unfolding Generations}

Comparison of manifold unfolding using original LLE, GLLE with EM algorithm, and GLLE with direct sampling is shown in Fig. \ref{figure_GLLE_generations}. 
For the simulations of this paper, we set $k=10$ to consider ten neighbors in $k$NN graph. 
This figure depicts four different generations of the proposed GLLE algorithms for each dataset. As can be seen, the GLLE embeddings are closely related to LLE embeddings. The GLLE generations have correctly unfolded the manifold with acceptable precision. The different generations of GLLE are also related because their reconstruction weights are sampled from the same probability distributions. The embedding of GLLE with EM algorithm for S-curve has become narrow which is interpretable; this is because of the most informative direction of the S-curve which is the narrow S-shape from its side. 
In all different generations of GLLE algorithms, the relations of neighbor points have been preserved as expected. 

\subsection{Analyzing the Effect of Covariance Scaling on GLLE}

Both the proposed GLLE versions work with the covariance matrix of reconstruction weights for every data point.   
We analyzed the impact of covariance scaling on the embedding generations of GLLE algorithms. Figure \ref{figure_GLLE_interpolation} illustrates several generations of GLLE algorithms with five different scales of covariance matrices of weights. We used the scaled covariance matrices $a \b{\Omega}_i$ and $a \b{\Gamma}_i$ with $\forall i \in \{1, \dots, n\}$ and $a \in \{0.01, 0.1, 1, 5, 10\}$ for GLLE with EM algorithm and GLLE with direct sampling, respectively. This figure shows that the learned covariance matrices in GLLE with EM algorithm and the derived covariance matrices for GLLE with direct sampling work properly because the manifolds have been correctly unfolded. As this figure shows, both GLLE versions have acceptable robustness to different scales of covariance matrices; although, the correct scale, i.e. $a=1$, often has the best performance as expected. Some large scales of covariance have resulted in high-variance and slightly less accurate embeddings, which is expected. 

\section{Conclusion}\label{section_conclusion}

In this paper, we proposed a generative version of LLE, named GLLE, where the linear reconstruction step in the input space is performed stochastically rather than deterministically. Two versions of stochastic linear reconstruction were proposed, one of which used EM and the other used direct sampling. The proposed GLLE was closely related to LLE, variational inference, factor analysis, and probabilistic PCA. Simulations on seevral datasets showed the effectiveness of the generated embeddings in comparison to LLE embedding.



\section*{Acknowledgment}

The authors thank and remember Prof. Sam T. Roweis (rest in peace) and Prof. Brendan Frey who proposed the initial raw idea of generative LLE (specifically GLLE with direct sampling), at the University of Toronto, with Prof. Ali Ghodsi years ago.

\bibliographystyle{IEEEtran}
\bibliography{references}

\end{document}